\documentclass{article}

\usepackage{arxiv}
\usepackage[utf8]{inputenc}
\usepackage[T1]{fontenc}
\usepackage{hyperref}
\usepackage{url}
\usepackage{booktabs}
\usepackage{amsfonts}
\usepackage{graphicx}
\usepackage{array}
\usepackage{tabularx}
\usepackage{microtype}
\graphicspath{{figures/}}

\title{Guideline-grounded Retrieval-augmented Generation for Ophthalmic Clinical Decision Support}

\author{
Shuying Chen \\
School of Artificial Intelligence and Data Science \\
University of International Business and Economics \\
Beijing, China \\
\And
Sen Cui \\
Tsinghua University \\
Beijing, China \\
\And
Zhong Cao\thanks{Corresponding author.} \\
Heidelberg Institute of Global Health, Faculty of Medicine and University Hospital \\
Heidelberg University \\
Heidelberg, Germany \\
\texttt{zhong.cao@uni-heidelberg.de}
}

\begin{document}
\maketitle

\begin{abstract}
In this work, we propose Oph-Guid-RAG, a multimodal visual RAG system for ophthalmology clinical question answering and decision support. We treat each guideline page as an independent evidence unit and directly retrieve page images, preserving tables, flowcharts, and layout information. We further design a controllable retrieval framework with routing and filtering, which selectively introduces external evidence and reduces noise. The system integrates query decomposition, query rewriting, retrieval, reranking, and multimodal reasoning, and provides traceable outputs with guideline page references.
We evaluate our method on HealthBench using a doctor-based scoring protocol. On the hard subset, our approach improves the overall score from 0.2969 to 0.3861 (+0.0892, +30.0\%) compared to GPT-5.2, and achieves higher accuracy, improving from 0.5956 to 0.6576 (+0.0620, +10.4\%). Compared to GPT-5.4, our method achieves a larger accuracy gain of +0.1289 (+24.4\%). These results show that our method is more effective on challenging cases that require precise, evidence-based reasoning. Ablation studies further show that reranking, routing, and retrieval design are critical for stable performance, especially under difficult settings.
Overall, we show how combining vision based retrieval with controllable reasoning can improve evidence grounding and robustness in clinical AI applications, while pointing out that further work is needed to be more complete .
\end{abstract}

\noindent\textbf{CCS Concepts:} Computing methodologies; Artificial intelligence; Information systems; Information retrieval; Applied computing; Life and medical sciences; Health informatics.

\noindent\textbf{Keywords:} Ophthalmology, Clinical Decision Support, RAG, Multimodal Retrieval, Visual Document Retrieval, HealthBench

\section{Introduction}
In the domain of ophthalmology, treatment decisions are highly based on clinical practice guidelines, which define treatment pathways, drug usage, follow-ups and referral conditions. The criticality in clinical practice could result in severe or even fatal consequences due to mistakes made during execution; consequently clinical decision making should rely on accurate, reproducible and auditable data for the sake of science and patients alike.

According to recent review results, ophthalmology is rapidly entering an era where both basic model and clinical verification are required. But privacy, bias and the quality of the generated clinical evidence are important issues \cite{ref1}. In this regard, there is an intense desire for decision-support that can be based on LLM/LMM; however, relying only on parametric knowledge brings in two significant dangers: firstly, outdated and missing knowledge causes unstable reactions toward the most recent evidenced based recommendation; second, it can hallucinate thresholds, grading, contraindications, or workflow for rare and/or complex cases.The real-world clinical validation of an ophthalmological use case shows how a structured agent decouples the vision processing, knowledge search, and diagnostics reasoning can help improve treatment plans, and avoid general purpose model fragility on rare diseases \cite{ref2}.

RAG is popularly used to address those problems: it retrieves outside information and uses the retrieved evidence context at generation time, the system enhances traceability and fact-consistency. Systematic review and meta-analysis show that RAG yields statistically significant gains over baselines on biomedical tasks, with actionable advice on how to deploy in the clinic \cite{ref3}. However, vanilla RAG assumes that we have good text extraction, chunking and vectorization.In reality, clinical guidelines that are filled with rich tables, flowcharts, complex layouts, embedded images, etc., which makes the process of text extraction and OCR pipeline brittle and expensive. Furthermore, even if the relevant evidence is retrieved, models might not use them properly because of incomplete contexts aggregation, leading to incomplete grounding or remaining hallucination \cite{ref4}. Visual document retrieval thus calls fora paradigm of directly embedding the page image for retrieval, avoiding complicated text extraction and taking full advantage of layout and graphical information, providing a new direction to retrievals based on guidelines.

We thus turn to the clinically focused problem of how to incorporate guideline evidence as visual pages in RAG for ophthalmic clinical QA and decision support, and how to design an agentic router that can make a controllable decision on when to use external retrieval versus direct generation, while ensuring traceability, robustness, and faithfulness, and reducing hallucinations.This paper makes four key contributions. 

First, we propose a page-level visual RAG method for ophthalmology guidelines. Each guideline page is used as an independent evidence unit. We directly retrieve page images instead of text, which avoids OCR errors and keeps important structures such as tables and flowcharts. 

Second, we design the retriever as an controllable retrieval system that has route and filter functions. It routes whether to retrieve or not, and filters out less relevant evidences. Thus it can balance factuality and fluency of responses. 

Third, we build an end-to-end multimodal pipeline. It includes query decomposition, query rewriting, retrieval, reranking, and answer generation.The system also returns guideline page URLs, so the outputs are traceable. 

Fourth, we evaluate our system on HealthBench with a doctor-based scoring protocol. The results show that our method improves performance on hard cases, especially in accuracy, and that controlled retrieval is important for stable and reliable output 

\section{Related Work}
 We categorize previous research into three categories: (i) foundation models and intelligent agents centered on ophthalmology, (ii) guideline-based retrieval-augmented generation (RAG) for clinical decision support across various specialties, and (iii) retrieval and evaluation methodologies for biomedical RAG systems. 

\subsection{Ophthalmology foundation models and intelligent agents}
 Multimodal Foundation Models and Domain-Specific Agents Recent work in computer vision for eye diseases have been focusing on multimodal foundation models and the domain-specific agents.Zhuang et al. (Zhuang, 2025) proposes a moduler reasoner to integrate visual analysis, retrieval and diagnosis using concrete improvements in the practical use cases for ophthalmic treatment \cite{ref2}. Sevgi, F. A.(2024), Personalized Instruction-Tailored and RetrievalAugmented GPT Models for Eye Education and Clinical Assistance, for the purposes of privacy, accountability, and other operational concerns \cite{ref5}. An extensive road map of vision-language foundation models for ophthalmology is provided in Chia(2024), observing continued problems with data bias, regulation and appropriate clinical validation. Apart from these changes, making everything simple and understandable is another major focus now \cite{ref1}.Han Wang (2025) combines knowledge graphs and contrastive learning together with organised “ clinical profile ” metrics to explain how ophthalmic AI works. Training \& Simulation \cite{ref6}. Luo(2025) proposes a retrieval augmented digital patient for improving taking an ophthalmic history showing that RAG-based approaches are worth learning \cite{ref7}. Grzybowski(2024) discusses AI algorithm to analyse the fundus images of ocular and systemic disease \cite{ref8} while Ong A.Y.(2025) analyses regulator-sanctioned ocular image-analysis AI aMD’s shortcoming on reporting norms and empirical data \cite{ref9}. Also, there is continuous development of disease specific DSS such as the staging algorithm for keratogenous developed by Muhsin et al.(2024) \cite{ref10}. 

\subsection{Guideline-grounded RAG for clinical decision support}
 Our concern is not just that the model could give us an answer, but that it gives an answer consistent with what clinicians would say. In this regard, Kresevic (2024) shows how systematic guideline reformulation together with RAG and agile prototyping improve precise guidance understanding, suggesting that both evidence format as well as search strategy matter for trustworthiness \cite{ref11}. Ong C.S.(2024). Proposes Surgery LLM that attempts to unify workflow and share the same goal \cite{ref12}. Bring evidence surgical guideline in peri operative assistant, showcasing application of guideline grounding to a real-world clinical use case, and Ke (2025) takes this idea further still with a comparison of RAG across 10 large LMs for preoperative risk assessment, reported better diagnostics and less hallucination \cite{ref13}.This supports the notion that structured retrieval can serve as a stabilizing constraint rather than merely an optional enhancement. The same pattern is important to note: Miao (2024) uses KDIGO-based corpora to apply RAG in nephrology \cite{ref14}, and Wada, 2025), as it allows for robust local deployment of radiology consultative AI while simultaneously enabling privacy preserving infrastructure—implementation considerations which have direct impact on practicality \cite{ref15}. Analogously, Chen (2025) combines RAG and leastto-most prompting strategy for treating low back pain \cite{ref16}; and Zhou (2024) builds GastroBot, a Chinese gastroenterology-based chatbot built upon carefully chosen guideline corpora \cite{ref17}.LLM responses are therefore more consistent, more clinically relevant when the input information is collected in a systematic manner and tied back to domain guidelines. and are less prone to hallucination, which makes it increasingly become a design principle rather than a performance hack. 
\subsection{Retrieval architecture and evaluation methodology}
 The research methodology of the field aim at solving below three problems: accuracy, resilience and rigor of assessment. Liu et al.(2025) places the development of Biomedical RAG system within existing clinic deployment standards through an extensive survey and meta analysis \cite{ref3}. To directly address evidence alignment, Prabha(2025) introduces structured, iterative self-query retrieval utilizing PICOT/SPICE frameworks, such as incorporating it in existing clinical decision making frameworks \cite{ref18}.Lopez et al.(2025) propose CLEAR, a method to incorporate entities into the retrieval process with lower token usage, slower response time but improved span recovery. During the design of searchers, this approach considers efficiency \cite{ref19}. Zhang (2025) studies “lost-in-the-middle” problem in medical question answering, and evaluate Brief Context reordering solutions that can ease out of context dilution with longer contexts \cite{ref20}. Gilbert (2024) call for stronger integration of knowledge graphs as a way to reduce loss of hallucination when curating medical information \cite{ref21}, and Gilani (2025) that propose CDE-Mapper: combination of retrieval and rule based mapping for mapping clinical items to CVs \cite{ref22}.Secondly from the perspective of system maintenance, Borchert(2025)argues that accurate retrieval pipes are needed to make timely revisions in guidelines possible and calculates the time lagbetween the appearance of a piece of new evidence and its resulting revision of a guideline.Reliance on time-synchronization serve as part of the integrity \cite{ref23}.
 
 Recent work proposes some benchmarks to evaluate medical RAGs: MIRAGE is a large scale evaluation benchmark which systematically compares different combinations of retrievers,corpora, and backbone models across multiple medical QA datasets to highlight the significance of retrieval design in terms of increasing accuracy \cite{ref24}. RAGCare-QA further provides a structured dataset to evaluate the RAG pipeline over theoretical medical knowledge with different levels of complexities in questions as well as retrievals required \cite{ref25}.However,these benchmarks mostly target static QA tasks and evaluate based on correctness alone, while missing any domain-specific metrics or task characteristics which are representative of clinical practice .in particular they do not directly assess dimensions such as safety, context awareness or adherence to the process of clinical reasoning. In order to overcome these shortcomings,we leverage the doctor-rubric based benchmark, HealthBench \cite{ref26} to evaluate health-care conversational agents in clinically realistic scenarios. HealthBench evaluates the model response from multiple aspects such as accuracy, completeness, context awareness, and communication quality, offering a more complete evaluation on the clinical utility and safety.Compared with previous QA-based benchmarks, HealthBench more closely matches the needs of real-world CDS, where responses are required to be not just right but also safe, contextualized, and clinically-relevant. 

\section{Method}

\subsection{Terminology}
 For clarity, we define key terms and abbreviations used throughout the paper. Retrieval-Augmented Generation (RAG) refers to the paradigm of augmenting language models with external knowledge retrieved at inference time. Question answering (QA) denotes the task of generating answers to user queries in a clinical context. DIRECT refers to the pathway where the model generates answers without external retrieval. TOS (object storage service) is used to store page-level guideline images and their metadata. FAISS is employed as the vector indexing library for efficient similarity search. ColQwen2.5 is used as the multimodal retriever to encode both queries and document images into a shared embedding space.
 
 To avoid terminological confusion, throughout this paper we consistently employ page level evidence units and page level retrieval terminology. We consider each individual guideline page image resulting from a pdf-to-image conversion as our atomic evidence unit for retrieval and citation, and call the retrieval module a page level visual retriever or visual page retriever, where the retrieved items are called evidence pages. Multimodal reasoning in this paper specifically means jointly feeding the textual query and guideline page image evidence to a vision language model for evidence aware response generation, and it excludes automatic diagnosis/inference over the patient’s OCT, fundus photograph, etc. clinical images.Here we refer to the clinical decision support as a question-answering and recommendation draft assistant which can help clinicians or trainees quickly find out guidelines’ evidences and generate traceable explanation recommendations, and it’s not a medical device that autonomously diagnoses or independently prescribes: any output should be used in conjunction with appropriate professional supervision and local clinical governance. 

\subsection{System architecture overview}
 We present the four-stage multimodal visual RAG (Retrieval- Augmented Generation) system for ophthalmology question answering (QA). named Oph-Guid-RAG for supporting controllable and traceable guideline-based generation with the constraint of clinical safety.

\begin{figure*}[t]
  \centering
  \includegraphics[width=\textwidth]{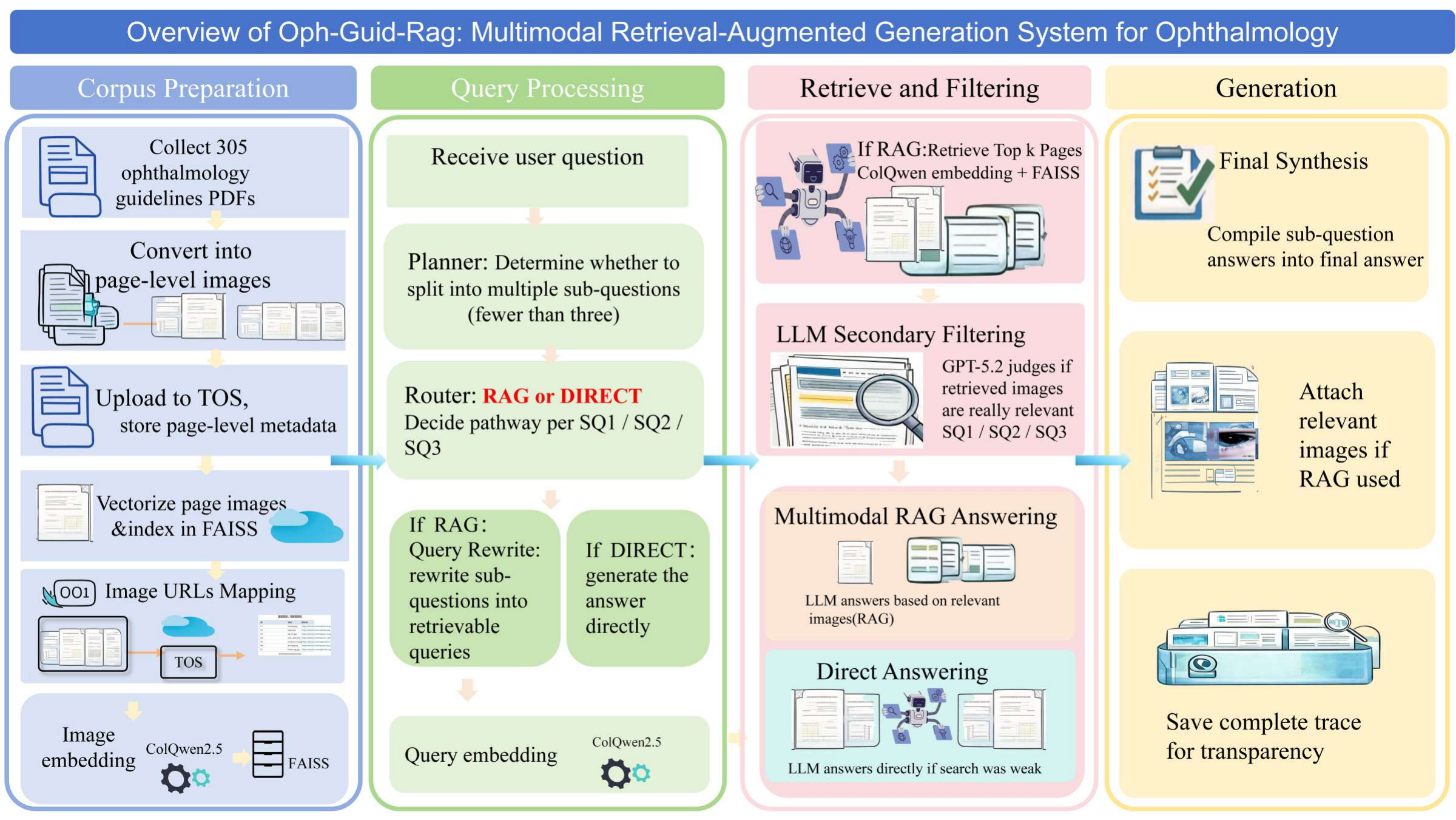}
  \caption{System Architecture of Guideline-Grounded Multimodal RAG for Ophthalmic Clinical Decision Support.}
  \label{fig:architecture}
\end{figure*}

 Specifically, we first prepare the off-line corpus by converting 305 ophthalmology guidelines from pdf to page level images, standardized into a uniform dimension (5390$\times$7940 pixel) , uploaded to TOS(object storage service), recorded with page level meta information and url mapping relation . The resulted page images are then encoded by ColQwen2.5 and indexed by FAISS as the vision knowledge base. Second, in the query processing stage, each incoming user queryis analyzed by a Planner, which optionally decomposes complex queries into up to three focused subquestions, denotedas SQ1, SQ2, and SQ3. A Router sends every subquestion down one of a RAGor a DIRECT branch. Subquestions passed into RAG go through a Query Rewrite module which produces one or two rewritings more suitable for retrieval on top of the guidelines corpus.Subquestions routed to DIRECT bypass retrieval and are answered directly by the model. Third, during retrieve-and-filter step, the rewritten queries are encoded by ColQwen2.5 and search over the FAISS index to get the top-k candidate page images and filter them by GPT-5.2, which measures their relevance with respect to both, the initial question as well as the generated subquestion. Only relevant pages are kept as valid evidence. When there’s not enough relevant evidence left, it goes back to the DIRECT path. Fourth, in the generation stage, the system carries out either evidence-grounded multimodal answer generation or direct answer generation with safety-oriented medical response constraints. A Final Synthesis module finally synthesizes all subanswers into one final answer while attaching referenced guideline images (if using RAG) and storing a complete process trace for audit ability.An overview of the proposed Oph-Guid-RAG framework is illustrated in Figure~\ref{fig:architecture}.

\subsection{Stage I: Corpus Preparation}
 In the offline corpus preparation stage, we got PDFs of ophthalmology-related guidelines from the Medlive (Yimaitong) platform during the ingestion stage. There were no failures when processing 305 PDF files. There are 7001 pages in the corpus, and each guideline document has an average of 22.95 pages.\begin{table*}[t]
\caption{Guideline Data Processing Statistics.}
\label{tab:table1}
\centering
\begin{tabular}{ll}
\toprule
Category & Statistic \\
\midrule
Number of Guideline Documents & 305 PDFs (0 failures) \\
Total Number of Pages & 7001 \\
Average Pages per Document & 22.95 \\
Rendering DPI & 720 DPI \\
Average Raw Image Resolution & $5908 \times 8063$ pixels \\
Processed Image Resolution & $5390 \times 7940$ pixels (resize + center padding) \\
Preprocessing Operations & No OCR/denoising/cropping/rotation \\
\bottomrule
\end{tabular}
\end{table*}
The guideline sources span a broad range of authorities and can be grouped into four main categories: 
 
(1)Global Authorities : This category consists of renowned international institutions such as the WHO , the American Academy of Ophthalmology (AAO).

(2) Government / National Organizations: This includes official national health and medical regulatory bodies.

(3) Provincial/Society Guidelines: Includes medical societies at the regional level and consensus documents at the provincial level. 

(4) Other Expert Consensus: This includes other statements of agreement from specialized experts. 

The mean page size for the guideline PDFs was 5908 x 8063 pixels, and we rendered every page to a high resolution image at 720 DPI. Post-rendering, each page is rescaled to be fitted into a target canvas of size 5390$\times$7940 pixels preserving its original aspect-ratio and then padded with a central white background so as to ensure matching input dimensionality for subsequent embeddings. No OCR was performed, automatic noise reduction, rotation and/or border trimming. We intentionally did not do any text extraction or structure parsing in order to preserve the original layout information like tables, flow charts, staging criteria, dosing threshold etc).In this approach every page is treated like an independent document and the retriever deals with precise images of original documents. We stored the cleaned up page image files into the object store and generated a JSON manifest file containing URLs for ease of use during the indexing and online inference phases. Our page-level visual indexing approach eliminates the need for standard text segmentation:which may split clinically relevant information in paragraphs, and destroy the document’s formatting. 

To support page-level visual retrieval, we adopt ColQwen2.5 as the visual document retriever. Both guideline page images and user-side retrieval queries are mapped into a shared vector space by ColQwen2.5, with page images encoded through the image encoder and textual queries encoded through the query encoder.The retriever outputs sequence-level embeddings, which we reduce to one fixed-length representation per image or query by mean pooling over the sequence dimension. This pooling strategy simplifies system integration and stabilizes downstream indexing and retrieval. 

In our present setup, we do not perform any further L2 normalization of the pooled embeddings; instead we construct a FAISS exact nearest-neighbor index (IndexFlatL2) and use the raw L2 distance as the retrieval score. This design preserves transparency in the retrieval process and makes the retrieval-confidence threshold easier to calibrate during inference. As a result, the offline stage produces a page-level visual knowledge base consisting of normalized page images, aligned metadata and URL mappings, and a FAISS-based ColQwen2.5 visual index for later retrieval. 

\subsection{Stage II: Query Processing}
 In the online inference stage, the system first receives the user query and processes it through a dedicated query processing module.Since real-world medical questions are often multi-intent, under specified, or contain multiple clinical constraints, the system begins with a Planner that determines whether the original query should remain intact or be decomposed into smaller focused subquestions. If decomposition is needed, the question is split into at most three subquestions, which allows downstream retrieval and answer generation to operate on more localized and semantically coherent units. 
 
 On each of these resulting subquestions, we apply a Router to decide if the subquestion will take a RAG pathway or a DIRECT pathway. The RAG pathway is meant for those questions which are dependent on evidences from guidelines, such as medication regimens, threshold-dependent criteria, contraindications, followup intervals, and structured management recommendations.In contrast, the DIRECT pathway is used for cases in which retrieval is unnecessary or unlikely to improve answer quality.
 
 For subquestions routed to the RAG pathway, the system further applies a Query Rewrite module that reformulates the subquestion into retrieval-oriented queries better aligned with the underlying guideline pages. These rewritten queries are designed to emphasize clinically meaningful entities or constraints, such as disease names, examination parameters, drug names, phases, cutpoints, or therapies. When a subquestion gets passed to DIRECT, the retrieval phase is skipped and the subquestion goes straight into answer extraction. 

\subsection{Stage III: Retrieval and Filtering}
 Visual Retrieval For every incoming subquestion to the RAG branch, we perform a visual search on the page-wise knowledge base.Rewritten retrieval queries are encoded by the ColQwen2.5 query encoder, while guideline pages have already been encoded offline by the image encoder into the same shared vector space. The resulting embeddings are indexed with FAISS, which is used to retrieve the nearest candidate evidence pages. In this way, retrieval operates directly overpage images rather than over extracted text passages, allowing the system to preserve page-level layout cues and visual evidence structure. 
 
 However, the retrieved candidates are not used directly. To reduce false-positive retrieval and improve evidence To enhance the precision, we further add one more LLMbased relevance filter layer on top of it. Specifically, we employ GPT-5.2 to examine each retrieved candidate page and determine whether it is genuinely relevant to both the original user query and the current focused subquestion. Only those pages judged as relevant are retained as valid evidence for downstream multimodal answering. 
 
 Once filtered, we decide if there is enough reason to believe in the evidence found or not. If so and there is enough relevant evidence then, the subquestion moves on to multimodal RAG answer. If there is no evidence survived after the filter step, or when a retrieval signal is judged to be too weak, the system triggers a retrieval quality fallback and reverts to the DIRECT pathway. This design prevents the generator from relying on semantically mismatched or low-confidence retrieval results, thereby improving robustness and reducing error propagation from the retriever to the final answer. 

\subsection{Stage IV: Generation and Final Synthesis}
 In the generation stage, the system answers each subquestion according to the pathway determined in the previous stage.If reliable evidence pages are available, the system performs multimodal RAG answering, in which the focused textual question and the retrieved page images are jointly provided to the generator so that the answer can be grounded in explicit guideline evidence. If reliable evidence is not available, the system instead performs DIRECT answering, relying on the generator’s general medical reasoning ability together with safety-oriented prompting. 
 
 After all subquestions have been answered, the system invokes a Final Synthesis module to merge the subanswers into one coherent final response. This synthesis step is responsible for preserving logical consistency, reducing redundancy across subanswers, and producing a fluent end-to-end answer to the original user query.When RAG evidence has been used, the system can additionally append the referenced image sources so that the final output remains traceable and auditable. 
 
 The system also records a structured process trace covering all major intermediate decisions, including question decomposition, routing results, rewritten queries, retrieved pages, filtering outcomes, final answer modes, and source references. This trace serves both as an auditing mechanism and as an analysis tool for later debugging, ablation studies, and system evaluation. Therefore, the final output of the system is not only an answer, but also includes the directly displayed guideline page images as visual evidence for verification and clinical interpretability, followed by a structured process trace that records how the answer was produced.

\begin{figure*}[t]
  \centering
  \includegraphics[width=\textwidth]{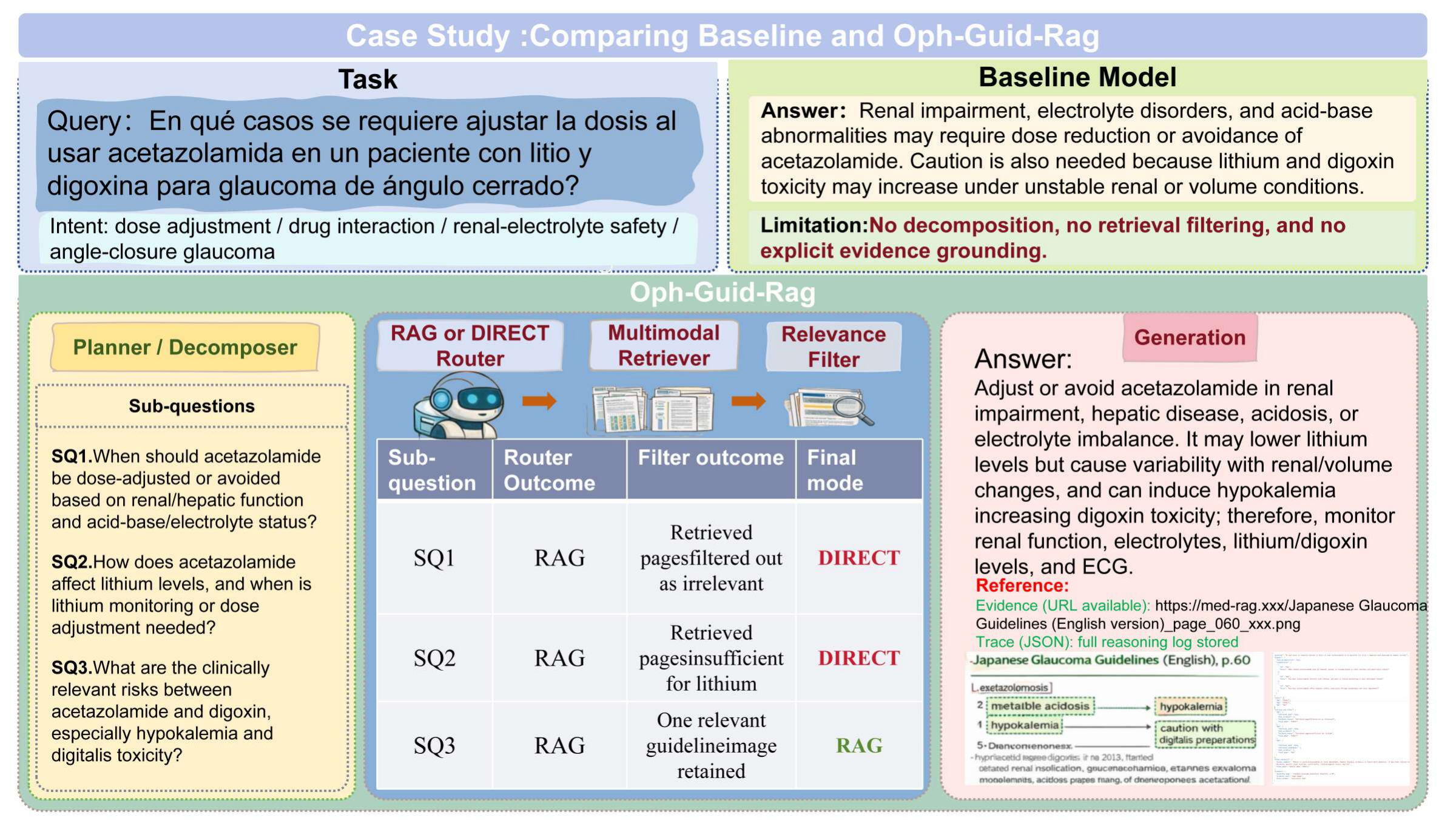}
  \caption{Case Study :Comparing Baseline and Oph-Guid-Rag}
  \label{fig:case-study}
\end{figure*}

To better illustrate how the proposed system operates in practice, we present a representative case study in Figure~\ref{fig:case-study}.
 Figure 2 shows a real example of how the system processes a clinically complex query. The input question involves multiple clinical considerations (e.g., dose adjustment, drug interaction, and electrolyte safety). The Planner first decomposes the query into three focused subquestions. Each subquestion is then routed to either the RAG or DIRECT pathway.For subquestions initially routed to RAG, the system retrieves candidate guideline pages and applies an LLMbased relevance filter. If no sufficiently relevant evidence is retained, the system falls back to the DIRECT mode. In this example, two subquestions fail to obtain reliable evidence after filtering and are therefore answered directly, while one subquestion retains a relevant guideline page and proceeds with multimodal RAG.Compared to the baseline model, which provides a general answer without explicit evidence grounding, the proposed system produces a more clinically grounded response.In addition to returning the supporting guideline page as visual evidence, the system also records and outputs a structured end-to-end process trace (trace.json), covering decomposition, routing decisions, rewritten queries, retrieval results, filtering results, and mode of final answer so that we have complete transparency, traceability and post-hoc verification of the reasoning process. 

\section{Evaluation}

\subsection{Dataset construction}
 The ophthalmology prompts we use in our work are taken from the full HealthBench dataset, as well as two of its official subsets. We use one, fully reproducible filtering rule to create the ophthalmology subset. We extracted for every example the question text from the first available text field (“question”,”prompt”, or ”content”), lowercased the string, and matched against an ophthalmology keyword list which was pre-defined by us.The key word list consisted of following words: ”ophthalmology,” ”eye,” ”retina,” ”glaucoma,” ”cataract,” ”cornea,” ”vision,” ”intraocular pressure,” ”fundus,” ”strabismus,” ”myopia,” ”hyperopia,” ”amblyopia,” ”macula,” ”vitreous,” and ”optic nerve.” If at least one keyword matched, an example was kept.

 To prevent a potential distributional bias induced through filtering on the subsets themselves, we used exactly the same deterministic rule for filtering the whole dataset and both official subsets; this ensures that the way in which we build up the ophthalmology subset is transparent, can be verified, and can be precisely reproduced. We collected in total 78 ophthalmology prompts: 16 were taken from the hard subset, and 62 from the consensus one.For easy verification and duplication, we release the precise filtering code and keywords list publicly on our GitHub repo6 . The final data split statistics are presented in Table~\ref{tab:table2}.
\begin{table}[t]
\caption{HealthBench Ophthalmic Subset Split Statistics.}
\label{tab:table2}
\centering
\begin{tabular}{lrr}
\toprule
Split & Number of Prompts & Percentage of Total \\
\midrule
Main & 78 & 100.00\% \\
Consensus & 62 & 79.49\% \\
Hard & 16 & 20.51\% \\
\bottomrule
\end{tabular}
\end{table}

\subsection{Metrics / Rubric}
 We use HealthBench that scores model output with rubrics (written by doctors) on a conversation-by-conversation basis and aggregates the rubric criteria into behavioral axes so that strengths and failure modes can be attributed in great detail \cite{ref1}.In our reporting, we follow the HealthBench axis definitions and focus on dimensions that have a direct impact on clinical usability and safety, such as: 
 
 (1) Accuracy: The clinical information and suggestions must be factually correct. 

 (2) Completeness: Are all the important parts of assessment and management covered? 

 (3) Context awareness: Does the system respond to situational cues and ask follow-up questions when information is missing? 
 
 (4) Communication quality: Is the tone appropriate for the audience? 
 
 (5) Follow-the-leader: adhering to the guidelines of both the user and the circumstance. 
 
 We consider the following key facets in our Hard-subset investigation: completeness and contextulization that demonstrate most clearly the rubric behaviors supporting safe and effective clinical conversation. All ratings were given as values between 0 and 1, where we take the mean. 

\subsection{Experimental setup and comparability}
 We test Oph-Guid-RAG against GPT-5.2, GPT-5.3, and GPT-5.4 to see how it compares. Please keep in mind that we made all of the baseline scores ourselves. We use the same pipeline for GPT- 5.2, GPT-5.3, and GPT-5.4, and we grade them all with the same HealthBench setup. This makes it possible to compare them directly and fairly in the same situation. 
 
 We also have ablation variants like no\_rerank, no\_query\_rewrite, and no\_router. To be fair, all models and all versions are tested in the same way. They use the same HealthBench conversation format, the same system safety prompt, the same decoding setting, and the same limit on the length of the output.The only thing that makes models different is how they get data back. Oph-Guid-RAG uses guideline page retrieval along with rerank, query rewrite, and routing. One of these modules is taken out by the ablation models. This retrieval pipeline is not used by the baseline models. There are no other changes. 
 
 We evaluate all output using the official HealthBench grading script, running their default model based grader. All runs share the same grading model, the same grading prompts and the same score aggregation logic. This guarantees that the score differences come only from model behavior and retrieval design, not from the evaluation process. We report results on both the full set and the hard subset. The hard subset contains more difficult questions and requires stronger reasoning and better use of evidence. We report overall score and five axes including accuracy, completeness, instruction following, context awareness and communication quality. 

\section{Experimental Results}

\subsection{Overall Performance on All Samples}
 We first compare our method with several strong baselines on the full evaluation set. The results are summarized in Table~\ref{tab:table3}.
\begin{table*}[t]
\caption{Performance Comparison on the All Set}
\label{tab:table3}
\centering\small
\resizebox{\textwidth}{!}{\begin{tabular}{lrrrrrr}
\toprule
Model & Overall Score & Accuracy & Completeness & Instruction Following & Context Awareness & Communication Quality \\
\midrule
Oph-Guid-Rag & 0.5524 & 0.6266 & 0.4614 & 0.3474 & 0.6212 & 0.4153 \\
GPT-5.2 & 0.5559 & 0.6317 & 0.501 & 0.5193 & 0.5027 & 0.447 \\
GPT-5.3 & 0.5484 & 0.569 & 0.4574 & 0.5672 & 0.5317 & 0.4717 \\
GPT-5.4 & 0.5856 & 0.6336 & 0.5314 & 0.4749 & 0.6333 & 0.5562 \\
\bottomrule
\end{tabular}}
\end{table*}

 The final score for our model is 0.55, similar as GPT-5.2(0.5559), but a little less than GPT-5.4 (0.5856). Our model obtains the accuracy of 0.6266, comparable with GPT-5.4 (0.6336), while better than GPT-5.3(0.5690) showing that adding guideline evidence helps to obtain a competitive correctness.Yet, the gap mainly comes from instruction following and communication quality. Our model obtains 0.3474 on instruction next, which is far lower than GPT- 5.2(0.5193), GPT-5.3(0.5672). Similar phenomenon appears in quality of communication, on which our score (0.4153) lags behind GPT-5.4’s (0.5562), suggesting that retrieval based generation is rigid to the response style that impacts on fluency and adherence to conversation instruction. Meanwhile, our model is strongly aware of the context(0.6212), better than that of GPT-5.2(0.5027) and GPT-5.3 (0.5317). This confirms that grounding on guideline pages improves the model’s ability to stay aligned with clinically relevant context. In general, the result shows that our approach has a balanced performance, being strong at accuracy and context grounding, leaving room for improving in instruction adherence and response naturalness. 

\subsection{Performance on Hard Subset}
 We further evaluate all models on a more challenging subset, where questions require deeper reasoning or precise guideline grounding. The results are shown in Table~\ref{tab:table4}.
\begin{table*}[t]
\caption{Performance Comparison on the Hard Set}
\label{tab:table4}
\centering\small
\resizebox{\textwidth}{!}{\begin{tabular}{lrrrrrr}
\toprule
Model & Overall Score & Accuracy & Completeness & Instruction Following & Context Awareness & Communication Quality \\
\midrule
Oph-Guid-Rag & 0.3861 & 0.6576 & 0.0483 & 0.1333 & 0.3969 & 0.9167 \\
GPT-5.2 & 0.2969 & 0.5956 & 0.0971 & 0.1333 & 0.3419 & 0.6167 \\
GPT-5.3 & 0.3344 & 0.5723 & 0.0183 & 0 & 0.2863 & 0.7833 \\
GPT-5.4 & 0.3756 & 0.5287 & 0.1139 & 0.1333 & 0.3901 & 0.8667 \\
\bottomrule
\end{tabular}}
\end{table*}

 On this subset we obtain the total score as 0.3861, which is better than that on GPT-5.2(0.2969), GPT-5.3 (0.3344), and slightly better than GPT-5.4 (0.3756). Against GPT-5.2 it’s an absolute improvement of +0.0892, indicating that our method is more robust under difficult conditions. The most significant accuracy advantage appears in accuracy. Our model is 0.6576, higher than GPT- 5.2(0.5956)+0.0620, GPT-5.4 (0.5287) by +0.1289. This shows that retrieval grounding is particularly beneficial when the task requires precise clinical knowledge. But the incompleteness is still a major limitation. We obtain the score of 0.0483, lower than GPT-5.4(0.1139) and GPT-5.2 (0.0971), showing that although we are able to discover right key points, does not always provide sufficiently detailed or fully developed answers. Interestingly, our model achieves a very high communication quality score (0.9167), better than GPT-5.4 (0.8667) and GPT-5.2 (0.6167). We conclude from this result that, after the model commits to an answer, it tends to provide coherent explanations. Taken together, these findings point towards the fact that trade-off. Our method improves factual correctness and robustness on hard cases, but at the cost of reduced completeness, likely due to conservative or partial use of retrieved evidence. 

\subsection{Ablation Study on All Samples}
 To better understand the contribution of each module, we conduct ablation experiments on the full dataset. The results are presented in Table~\ref{tab:table5}.
\begin{table*}[t]
\caption{Ablation Experiment Results of Different Model Configurations on All Subset}
\label{tab:table5}
\centering\small
\resizebox{\textwidth}{!}{\begin{tabular}{lrrrrrr}
\toprule
Model & Overall Score & Accuracy & Completeness & Instruction Following & Context Awareness & Communication Quality \\
\midrule
Oph-Guid-Rag & 0.5524 & 0.6191 & 0.4614 & 0.312 & 0.6212 & 0.4153 \\
no\_rerank & 0.5266 & 0.5816 & 0.4832 & 0.3354 & 0.5255 & 0.3799 \\
no\_query\_rewrite & 0.5574 & 0.6441 & 0.5505 & 0.4183 & 0.5267 & 0.3323 \\
no\_router & 0.55 & 0.6266 & 0.5088 & 0.3474 & 0.5851 & 0.402 \\
\bottomrule
\end{tabular}}
\end{table*}

 Removing the reranking module leads to a drop in overall score from 0.5524 to 0.5266 ($-$0.0258), and a decrease in accuracy from 0.6191 to 0.5816 ($-$0.0375).This confirms that reranking plays a critical role in improving retrieval quality and downstream correctness. Removing query rewriting results in an overall score of 0.5574, which is slightly higher than the full model. It also increases completeness from 0.4614 to 0.5505 (+0.0891). However, this comes at the cost of worse communications quality (0.3323 vs. 0.4153). This indicates that query rewriting introduces a bias towards more precise, but shorter answers, while removing it allows more verbose answers. Removing the router causes an overall score of 0.55, which is close to the full model. However, context awareness drops from 0.6212 to 0.5851 (-0.0361). These results indicate that routing enables selective application of retrieval where necessary to improve the contextual grounding. Overall, reranking is the most critical component on the full dataset, while the effects of query rewriting and routing are more subtle and task-dependent. 

\subsection{Ablation Study on Hard Subset}
 We further analyze the ablations on the hard subset, as shown in Table~\ref{tab:table6}.
\begin{table*}[t]
\caption{Ablation Experiment Results of Different Model Configurations on Hard Subset}
\label{tab:table6}
\centering\small
\resizebox{\textwidth}{!}{\begin{tabular}{lrrrrrr}
\toprule
Model & Overall Score & Accuracy & Completeness & Instruction Following & Context Awareness & Communication Quality \\
\midrule
Oph-Guid-Rag & 0.3861 & 0.6576 & 0.0483 & 0.1333 & 0.3969 & 0.9167 \\
no\_rerank & 0.2817 & 0.4461 & 0.0883 & 0 & 0.4026 & 0.4833 \\
no\_query\_rewrite & 0.3347 & 0.5654 & 0.1599 & 0.1333 & 0.4179 & 0.65 \\
no\_router & 0.3603 & 0.4835 & 0.066 & 0 & 0.4835 & 0.65 \\
\bottomrule
\end{tabular}}
\end{table*}

 The impact of reranking becomes much more significant under difficult conditions. Removing reranking reduces the overall score from 0.3861 to 0.2817 ($-$0.1044), and accuracy drops sharply from 0.6576 to 0.4461 ($-$0.2115). This shows the importance of selecting good evidence to answer complex clinical questions, while removing query rewriting decreases the overall score to 0.3347 ($-$0.0514), but increases completeness from 0.0483 to 0.1599 (+0.1116). This indicates that query rewriting improves precision but may overly constrain the retrieved evidence, resulting in partial answers. If we remove the router, then the overall score is 0.3603 ($-$0.0258). The accuracy drops significantly from 0.6576 to 0.4835 ($-$0.1741), while the context-awareness rises from 0.3969 to 0.4835 (+0.0866). This suggests that forcing all queries through retrieval adds more context, but also more noise, which hurts the correctness.In other words, these results show that on hard cases, reranking is essential, routing allows for trading precision against noise, while queryrewriting comes with a completeness-focus trade-off. 

\subsection{Discussion}
 The experimental results reveal that our method demonstrates clear advantages on the hard subset, which represents the most challenging scenarios in HealthBench \cite{ref26}.

 Following the official benchmark design, we use HealthBench Hard as a set of especially hard examples that current frontier models struggle to solve, and was specifically designed to challenge models at realistic levels of clinical reasoning. These cases are not simple factual questions. Instead, they are multi-turn clinical conversations that require reasoning under uncertainty, integration of context across turns, and adherence to detailed physician-defined evaluation criteria. 
 
 The difficulty of these problems lies in the fact that they require evidence-grounded reasoning under multiple constraints, not just memorization. General medical knowledge may not suffice to answer many questions and additional conformity to certain guideline statement including threshold values, contraindications, and treatment pathways. 
 
 At the same each response is scored on multiple clinical criteria (e.g., accuracy, completeness, safety, and requiring communication, where partial correctness is not sufficient. Both of these tasks are additionally made complex by uncertainty and missing data, for which the model should refrain from making overconfident statements and deliver contextualized recommendations. In these scenarios, typical LLMs are prone to failure since they rely on the parametric knowledge and often cannot meet multiple clinical requirements at once. Under this setting, our method shows consistent advantages by grounding responses in guideline evidence and carefully controlling the retrieval process. By retrieving page-level guideline content, the model can rely on authoritative sources instead of approximate internal knowledge, leading to clear gains in accuracy, with improvements of +0.0620 over GPT-5.2 and +0.1289 over GPT-5.4 on the hard subset.

 At the same time, routing and filtering help introduce external evidence only when necessary, avoiding irrelevant context, while reranking ensures that the most relevant guideline pages are selected. This design is critical on hard cases, as removing reranking leads to the largest drop ($-$0.1044 in overall score and $-$0.2115 in accuracy), and removing the router also causes a substantial accuracy decrease ($-$0.1741). Overall these results suggest our approach performs better in cases where there are clear evidential reasons to be able to reason precisely, while it continues to generate incomplete responses possibly because of its conservative use of evidence implying that better evidence aggregation should be explored in further research. 

\section{Limitations and Risks}
 Despite the strong empirical performance, several limitations remain. The system relies on the coverage and quality of the guideline corpus, which is inherently incomplete and may not capture recent updates, regional variations, or rare cases, leading to potential failures when relevant evidence cannot be retrieved. Meanwhile, the retrieval and the filtering pipeline comes with a precision-completeness trade-off, as reranking and LLM based filtering improve accuracy by removing noise, they may also discard partially relevant evidence, resulting in concise but incomplete replies (as seen from the hard subset results). The routing scheme can also be unreliable and sometimes take a wrong turn, either adding extraneous retrieval or omitting required evidence, both of which could impact factual grounding. Last but not least, we design our system to be used as a clinical decision support tool instead than a diagnostic system. Although it is encouraged to express uncertainty and avoid overconfident conclusions, it may still generate inappropriate or incomplete suggestions in some cases, and its outputs should therefore be interpreted with caution under professional supervision. 

\section{Conclusion}
 We introduce a multimodal visual RAG system for clinical decision support in the field of ophthalmology that is capable of enable evidence-grounded and controllable generation. By constructing a page-level guideline corpus and directly retrieving original guideline images as evidence, our approach avoids conventional text segmentation and preserves clinically relevant structures. In addition to the above, we propose a principled inference pipeline consisting of query-decomposition, The routing, query rewriting, retrieval, reranking and multimodal reasoning enable our model to flexibly leverage external knowledge on a selective basis during generation. 
 
 Extensive experiments on HealthBench demonstrate that our method achieves competitive overall performance and shows clear advantages on challenging cases. In particular, on the hard subset, our system outperforms strong baselines on both the accuracy and total scores, suggesting that it is effective at dealing with more complex, evidential clinical reasoning tasks. The ablation results further confirm that reranking, routing, and retrieval design all play critical roles in improving performance under difficult conditions.

 Meanwhile, we also show a key trade-off between accuracy and completeness: while grounding on guideline evidence increases factual correctness and context awareness, and can result in a less aggressive reaction as well and decreased coverage. It suggests that a future effort should be dedicated to improved evidence aggregation and further long generation policies while not damaging the factuality of inference. Overall, our paper provides a practical step toward trustworthy and traceable clinical AI systems, and demonstrates the potential of combining multimodal retrieval with structured reasoning for real-world medical applications. 

\section*{Acknowledgments}
 This study gratefully acknowledges the contributions of two researchers involved in data collection and preprocessing, as well as the computational support provided by the High-Performance Computing Platform of the Chinese Academy of Medical Sciences.

We also acknowledge the use of publicly available resources, including the OpenAI HealthBench dataset, Medlive, and guideline documents released by various institutions, which provided essential support for this research.

\end{document}